\pdfoutput=1
\documentclass[11pt,a4paper]{article}
\usepackage[hyperref]{emnlp2020}
\usepackage{times}
\usepackage{latexsym}

\usepackage{microtype}

\usepackage{url}

\aclfinalcopy 

\usepackage{amsmath}
\usepackage{tikz-dependency}
\usepackage{caption}
\usepackage{graphicx}
\usepackage{multirow}

\usepackage{algorithm}  
\usepackage{algorithmic}

\usepackage{array}
\newcommand{\PreserveBackslash}[1]{\let\temp=\\#1\let\\=\temp}
\newcolumntype{C}[1]{>{\PreserveBackslash\centering}p{#1}}
\newcolumntype{R}[1]{>{\PreserveBackslash\raggedleft}p{#1}}
\newcolumntype{L}[1]{>{\PreserveBackslash\raggedright}p{#1}}

\makeatletter
\def\blfootnote{\xdef\@thefnmark{}\@footnotetext}
\makeatother

\usepackage{color}

\title{Combining Self-Training and Self-Supervised Learning for Unsupervised Disfluency Detection}

\author{Shaolei Wang, Zhongyuan Wang, Wanxiang Che$^{*}$, Ting Liu \\
	Research Center for Social Computing and Information Retrieval,  \\ 
	Harbin Institute of Technology, Harbin, China  \\
	{\tt \{slwang, zywang, car, tliu\}@ir.hit.edu.cn} }

\date{}

\begin{document}
\maketitle
		\begin{abstract}
						
			Most existing approaches to disfluency detection heavily rely on human-annotated corpora, which is expensive to obtain in practice.			
			There have been several proposals to alleviate this issue with, for instance,  self-supervised learning techniques, but they still require  human-annotated corpora. 
			In this work, we explore the unsupervised learning paradigm which can potentially work with unlabeled text corpora that are cheaper and easier to obtain.
			Our model builds upon the recent work on Noisy Student Training,  a semi-supervised learning approach that extends the idea of self-training.
			Experimental results on the commonly used English Switchboard test set show that our approach achieves competitive performance compared to the previous state-of-the-art supervised systems using contextualized word embeddings (e.g. BERT and ELECTRA).

		\end{abstract}

		\section{Introduction}

			\blfootnote{
	
	\hspace{-0.65cm}  
	*Email corresponding.
}

\begin{figure}[t]
			\small
			\vspace{0.5em}
			\centering\includegraphics[width=70mm]{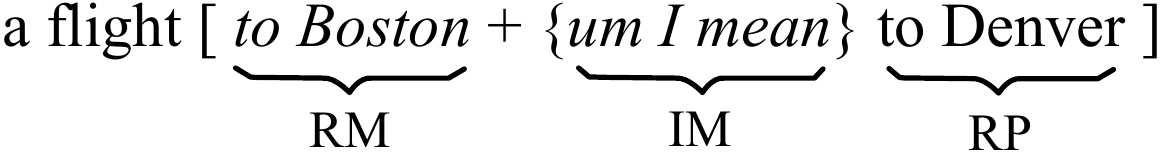}
			\vspace{-0.8em}
			\caption{A sentence from the English Switchboard corpus with disfluencies annotated. RM=Reparandum, IM=Interregnum, RP=Repair. The preceding RM is corrected by the following RP. }
			\label{fig:example}
			\vspace{-0.6em}
		\end{figure}

		Automatic speech recognition (ASR) outputs often contain various disfluencies, which  is a characteristic of spontaneous speech and create barriers to subsequent text processing tasks like parsing, machine translation, and summarization.
		Disfluency detection~\citep{zayats2016disfluency, wang-che-liu:2016:COLING, wu-EtAl:2015:ACL-IJCNLP} focuses on recognizing the disfluencies from ASR outputs. 
		As shown in Figure \ref{fig:example}, a standard annotation of the disfluency structure indicates the reparandum (words that the speaker intends to discard), the interruption point (denoted as `+', marking the end of the reparandum), an optional interregnum (filled pauses, discourse cue words, etc.) and the associated repair~\citep{shriberg1994preliminaries}.
		
		\begin{table}[t]\small
			\setlength{\tabcolsep}{5pt}
			
			\begin{center}
				\renewcommand{\arraystretch}{1.3}
				\begin{tabular}{|l|c|}
					\hline
					\bf    Type &  \bf Annotation \\
					\hline
					repair &  [ the + they 're ] voice activated\\ 
					\hline
					repair &  [ we want + \{well\} in our area we want ] to\\ 
					\hline
					repetition  &  [we got + \{uh\} we got ] to talking\\
					\hline
					restart & [ we would like + ] let's go to the\\
					\hline
				\end{tabular}
			\end{center}
			\vspace{-0.6em}
			\caption{ Different types of disfluencies.}
			\label{disf_example}
			\vspace{-0.6em}
		\end{table}
		
		Ignoring the interregnum, disfluencies are categorized into three types: restarts, repetitions and corrections. 
		Table \ref{disf_example} gives a few examples. 
		Interregnums are relatively easier to detect as they are often fixed phrases, e.g. ``uh'', ``you know". 
		On the other hand, reparandums are more difficult to detect in that they are in free form. 
		As a result, most previous disfluency detection work focuses on detecting reparandums.

		Most work~\citep{zayats2018robust, lou2018disfluency, wang2017transition, jamshid-lou-etal-2018-disfluency, zayats2019giving} on disfluency detection heavily relies on human-annotated corpora, which is scarce and expensive to obtain in practice. 
		There have been several proposals to alleviate this issue with, for instance,  self-supervised learning \citep{wang2019multi} and semi-supervised learning techniques \citep{C18-1299}, but they still require  human-annotated corpora. 
			In this work, we completely remove the need of human-annotated  corpora and propose a novel method to train a  disfluency detection system in a completely unsupervised manner, relying on nothing but unlabeled text corpora.
			
			Our model builds upon the recent work on Noisy Student Training \citep{xie2019self},  a semi-supervised learning approach based on  the idea of self-training.
			Noisy Student Training first trains a supervised model on labeled corpora and uses it as a teacher to generate pseudo labels for unlabeled corpora. 
			It then trains a larger model as a student model on the combination of labeled and pseudo labeled corpora. 
			This process is iterated  by putting back the student as the teacher.
			The result showed that it is possible to use unlabeled corpora to significantly advance both accuracy and robustness of state-of-the-art supervised models. 
			However, the performance of Noisy Student Training still relies on  human-annotated corpora. 
			
\begin{figure}[t]
			\small
			\vspace{0.5em}
			\centering\includegraphics[width=75mm]{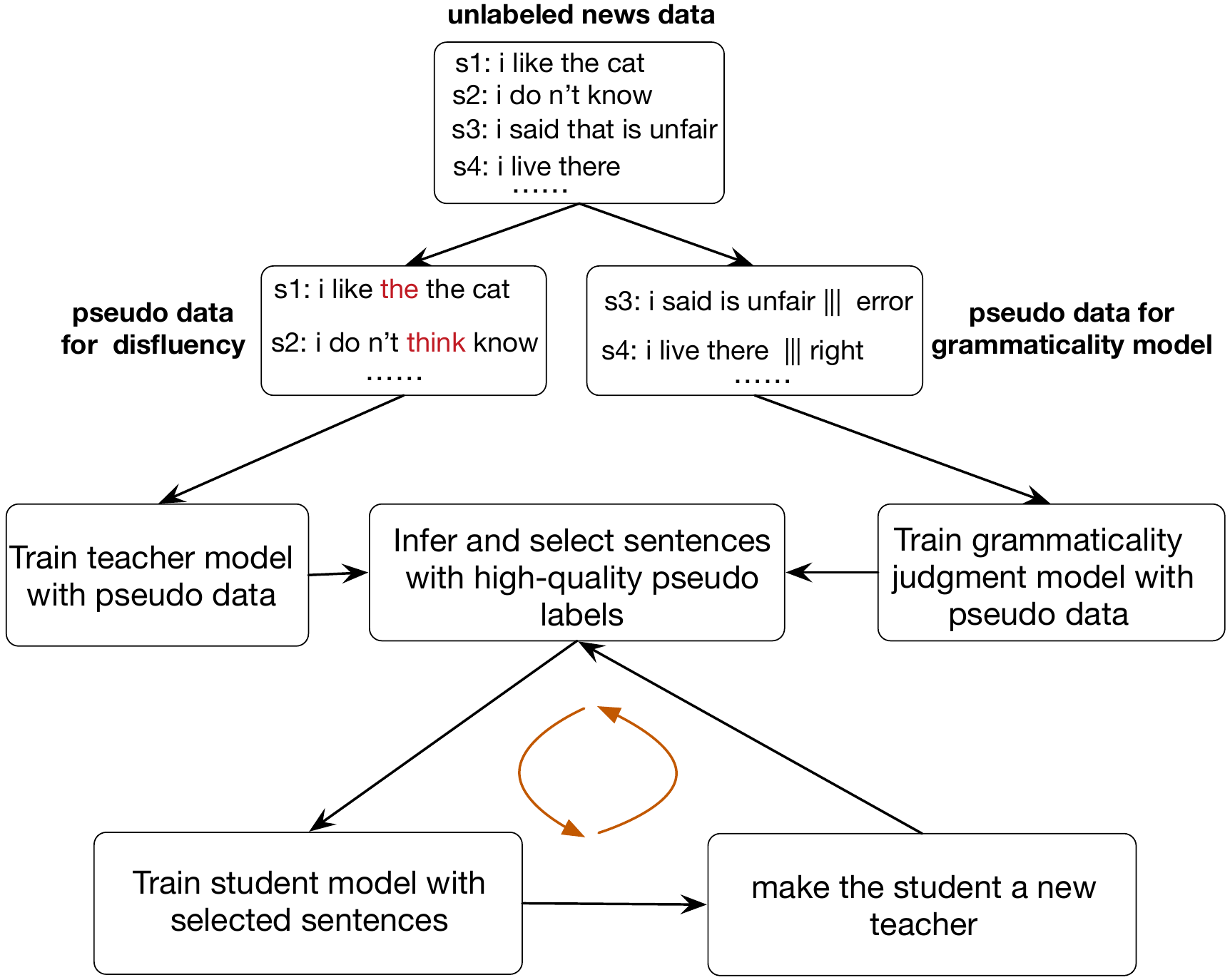}
			\vspace{-0.4em}
			\caption{Illustration of our proposed methods. }
			\label{network}
			\vspace{-0.4em}
		\end{figure}

			In this work,  we extend Noisy Student Training to unsupervised disfluency detection by  combining self-training and self-supervised learning methods.
			More concretely,  as shown in Figure \ref{network},  we use the self-supervised learning method to train a  weak disfluency detection model on large-scale pseudo training corpora as a teacher, which completely remove the need of human-annotated  corpora. 
			We also use the self-supervised learning method to train a sentence grammaticality judgment model to help select sentences with high-quality pseudo labels.

			Experimental results on the commonly used English Switchboard set show that our approach achieves competitive performance compared to the previous state-of-the-art supervised systems using contextualized word embeddings (e.g. BERT and ELECTRA).  
			Besides the experiment on the commonly used English Switchboard set, we  evaluate our approach  on another three different speech genres, and also achieve  competitive performance compared to the supervised systems using contextualized word embeddings.
			
			The code is released\footnote{https://github.com/scir-zywang/self-training-self-supervised-disfluency/}.

\begin{algorithm}[t]\small
	\caption{: Learning algorithm of our unsupervised model for disfluency detection}\label{algo:blah}
	\begin{algorithmic}[1]
		\REQUIRE Pseudo data for disfluency detection $(x_i, y_i)_{i=1}^N$, pseudo data for grammaticality judgment model $(\hat{x}_i, \hat{y}_i)_{i=1}^M$,  and unlabeled ASR outputs $\{\tilde{x}_1,\tilde{x}_2, ..., \tilde{x}_K\}$.
	\STATE Learn  sentence grammaticality judgment model  $\theta^g_*$ which minimizes the cross entropy loss on $(\hat{x}_i, \hat{y}_i)_{i=1}^M$
	\begin{equation}\nonumber
	\frac{1}{M} \sum_{i=1}^{M}\ell(\hat{y}_i, f^g(\hat{x}_i, \theta^g))
	\end{equation}
		\STATE Learn  teacher  model  $\theta^t_*$ which minimizes the cross entropy loss on $(x_i, y_i)_{i=1}^N$
				\begin{equation}\nonumber
					\frac{1}{N} \sum_{i=1}^{N}\ell(y_i, f^t(x_i, \theta^t))
				\end{equation}
				
		\STATE Use teacher model to generate pseudo labels for $\{\tilde{x}_1,\tilde{x}_2, ..., \tilde{x}_L\}$ collected from $\{\tilde{x}_1,\tilde{x}_2, ..., \tilde{x}_K\}$ by random sampling
						\begin{equation}\nonumber
						\tilde{y}_i = f^t(\tilde{x}_i, \theta^t_*), \forall{i} = 1, ... , L
						\end{equation}
		\STATE Use  sentence grammaticality judgment model to help select sentences with high-quality  pseudo labels $(\tilde{x}_i, \tilde{y}_i)_{i=1}^J$ from $(\tilde{x}_i, \tilde{y}_i)_{i=1}^L$.
		\STATE Learn a student model $\theta^s_*$ which minimizes the cross entropy loss on $(\tilde{x}_i, \tilde{y}_i)_{i=1}^J$
		
				\begin{equation}\nonumber
				\frac{1}{J} \sum_{i=1}^{J}\ell(\tilde{y}_i, f^s(\tilde{x}_i, \theta^s))
				\end{equation}
		
		\STATE Iterative training until the performance stops growing: Use the student as a teacher and go back to step 3.

	\end{algorithmic}
\end{algorithm}

\section{Proposed Approach}

\subsection{Unsupervised Training Process}

Algorithm \ref{algo:blah} and Figure \ref{network} give an overview of our unsupervised training. 
The inputs to the algorithm are all unlabeled sentences,  including news data and ASR outputs. 
We first construct large-scale pseudo data by randomly adding or deleting words to a fluent sentence, and use the self-supervised learning method to train a  sentence grammaticality judgment model. 
The sentence grammaticality judgment model has the ability to judge whether an input sentence is grammatically-correct or not.
We then construct large-scale pseudo data by randomly adding words to a fluent sentence, and use the self-supervised learning method to train a  weak disfluency detection model as a teacher. 
Next, we use the teacher model to generate pseudo labels on unlabeled ASR outputs.
Once a sentence is given correct  pseudo  labels, the rest  after deleting the  words with disfluency labels  is fluent and grammatically-correct.
Based on this fact,  we use the  sentence grammaticality judgment model  to help select sentences with high-quality pseudo labels.
We then train a student model on the selected  pseudo labeled sentences. 
 Finally, we iterate the process until performance stops growing by putting back the student as a teacher to generate new pseudo labels and train a new student. 
 We choose the student model achieving the best performance on human-annotated  dev set as our final model.

\subsection{System Architecture}
\subsubsection*{Train Teacher Model}
Traditional self-training  method trains the teacher model on labeled corpus.
In our work, we completely remove the need of human-annotated  corpora and use the self-supervised learning method to train a  weak disfluency detection model as a teacher.

We first construct large-scale pseudo data for the teacher model inspired by the work of \citet{wang2019multi}.
Let $S$ be an ordered sequence, which is taken from raw unlabeled news data, assumed to be fluent. 
We start from  $S$ and introduce random perturbations to generate a disfluent sentence $S_{disf}$. 
More specifically, we propose two types of perturbations:
\begin{itemize}
	\item \textit{Repetition$(k)$} : the $m$ (randomly selected from $one$ to $six$) words starting from the position $k$ are repeated.
	\item \textit{Inserting$(k)$} : we randomly pick a $m$-gram ($m$ is randomly selected from $one$ to $six$) from the news corpus and insert it to the position $k$.
\end{itemize}
For $S$, we randomly choose $one$ to $three$ positions, and then randomly take one of the two perturbations for each selected position to generate the disfluent sentence $S_{disf} = \{w_1, w_2, . . . , w_n\}$.
The training goal is to detect the added noisy words by associating a label for each word, where the labels $D$ and $O$ means that the word is an added word and a fluent word, respectively.
We directly  fine-tune the ELECTRA model (the discriminator) \citep{clark2020electra} on our pseudo  data.
Note that  the distribution of our pseudo data is different from the distribution of the gold disfluency detection data, which limits the performance of our teacher model on real test data.

\subsubsection*{Grammaticality Judgment Model}

Once a sentence $\{w_1, w_2, . . . , w_n\}$  is given correct  pseudo  labels $\{t_1, t_2, . . . , t_n\}$ by a teacher model,  the rest parts $\{\bar{w}_1, \bar{w}_2, . . . , \bar{w}_m\}$ by deleting the words with label $D$ is fluent and grammatically-correct.
Based on this fact,  we train a  sentence grammaticality judgment model  to help select sentences with high-quality pseudo labels.

We first construct large-scale pseudo data for the sentence grammaticality judgment model.
The input contains two kinds of sentences: 	(i) $S_{right}$ which is directly  taken from raw unlabeled news data.
(ii)  $S_{error}$ which is generated by adding some perturbations to $S_{right}$.
We  introduce  three types of perturbations to generate $S_{error}$. 
The first two types of perturbations are  \textit{Repetition$(k)$} and \textit{Inserting$(k)$} as described previously.
The third type of perturbations is:
\begin{itemize}
	\item \textit{Delete$(k)$} : for selected position $k$, $m$ (randomly selected from $one$ to $six$)  words starting from this position are deleted.
\end{itemize}
For an input sentence $S$, we randomly choose $one$ to $three$ positions, and then randomly take one of the three perturbations for each selected position to generate the disfluent sentence $S_{disf} = \{w_1, w_2, . . . , w_n\}$.
The training goal is to detect the type of an input sentence, where the labels $right$ and $error$ means that the sentence is grammatically-correct and grammatically-incorrect, respectively.
We  directly fine-tune the ELECTRA model (the discriminator) \citep{clark2020electra} on our pseudo  data.

\subsubsection*{Infer and Select Sentences}
We use the teacher model to generate pseudo labels on unlabeled ASR outputs.
The performance of  teacher model starts at a very low level, and it will bring too much noise
if we directly use the full unlabeled ASR outputs.
So we gradually increase the amount of unlabeled ASR outputs by random sampling from the full unlabeled ASR outputs in each iteration.

For an input sentence $S = \{w_1, w_2, . . . , w_n\}$,  the teacher model give  pseudo  labels  $T = \{t_1, t_2, . . . , t_n\},  \forall{t_i} \in{\{O, D\}}$.
Limited by the performance of  teacher model, it will bring  much noise if we directly train a student model on all the selected  pseudo labeled sentences. 
We use the sentence grammaticality judgment model  to help select sentences with high-quality pseudo labels.
Given a sentence $S = \{w_1, w_2, . . . , w_n\}$ and its pseudo labeles $T = \{t_1, t_2, . . . , t_n\}$, we get a sub-sentence $S_{sub} =\{\bar{w}_1, \bar{w}_2, . . . , \bar{w}_m\}$ by deleting the words with the label $D$.
If the sentence grammaticality judgment model   generates  $right$ label on  $S_{sub}$, we assume that  the pseudo labels $T$ is the same as gold labels and keep $(S, D)$ for student model training.

\subsubsection*{Train Student Model}
In this step, we directly fine-tune the first  teacher model as shown in Step 2 of Algorithm \ref{algo:blah}   on the selected  pseudo labeled ASR outputs, instead of fine-tuning the ELECTRA model.
Although the difference of distribution between our pseudo data and the golden disfluency detection data  limits the performance of  teacher model,  this stage converges faster than  fine-tuning the ELECTRA model as it only needs to adapt to the idiosyncrasies of the target disfluency detection data.

\section{Experiment}

\subsection{Settings}
\textbf{Dataset}.\quad   
English Switchboard (SWBD) \citep{godfrey1992switchboard} is the standard and largest ($1.73 \times 10^5$ sentences for training ) corpus used for disfluency detection. 
We use  English Switchboard as  main  data.
Following the experiment settings in \newcite{charniak2001edit}, we split the Switchboard corpus into train, dev and test set as follows: train data consists of all sw[23]$*$.dff files, dev data consists of all sw4[5-9]$*$.dff files and test data consists of all sw4[0-1]$*$.dff files.
Following \newcite{honnibal2014joint}, we lower-case the text and remove all  punctuations and partial words.\footnote{words are recognized as partial words if they are tagged as `XX' or end with `-'.}
We also discard the `um' and `uh' tokens and merge `you know' and `i mean' into single tokens.

In addition to Switchboard, we test our models on three out-of-domain publicly available datasets annotated with disfluencies \citep{zayats2014multi, zayats2018robust}:
	\begin{itemize}
		\item \textbf{CallHome:} \quad phone conversations between family members and close friends;
		\item \textbf{SCOTUS:} \quad transcribed Supreme Court oral arguments between justices and advocates;
		\item \textbf{FCIC:} \quad two transcribed hearings from Financial Crisis Inquiry Commission.
	\end{itemize}
The size of training and test sets for all corpora are given
in Table \ref{Corpus}.
	
Unlabeled sentences include news data and ASR outputs. 
News data are randomly extracted from WMT2017 monolingual language model training data (News Discussions. Version 2).\footnote{http://www.statmt.org/wmt17/translation-task.html}
Then we use the methods described in Section 2.2 to construct the pre-training dataset for the teacher model and grammaticality judgment model.
The training set of the teacher model contains 2 million sentences.
We use 5 million sentences for the grammaticality judgment model, in which half of them are grammatically-incorrect sentences and others are grammatically-correct sentences directly extracted from the news corpus.
The unlabeled ASR outputs we use include Fisher Speech Transcripts Part 1 \citep{cieri2004fisher} and Part 2 \citep{cieri2005fisher}, which contains about 835k sentences.

\noindent	\textbf{Metric}.\quad Following previous works \citep{ferguson-durrett-klein:2015:NAACL-HLT}, token-based precision (P), recall (R), and F1 are used as the evaluation metrics.

\begin{table}[t]\small
	\setlength{\tabcolsep}{13pt}
	
	\begin{center}
		\renewcommand{\arraystretch}{1.3}
		\begin{tabular}{l|c|c}
			\hline
			\bf    Corpora & \bf training & \bf test \\
			\hline
			SWBD & 1.3M  & 65K \\ 
			SCOTUS & 46K & 30K\\ 
			CallHome & - & 43K\\ 
			FCIC & - & 54K\\ 
			\hline
			
		\end{tabular}
	\end{center}
	\caption{ The number of words in training and testing data for
		different corpora. Note that we do not use the training data for our unsupervised methods.}
	\label{Corpus}
\end{table}

\subsection{Training Details}
In all experiments  including the ELECTRA model, we use English  ELECTRA-Base model with 110M  hidden units, 12 heads,  12 hidden layers.\footnote{https://github.com/google-research/electra}
For the  self-supervised teacher models  and grammaticality judgment model, 
we use streams of 128 tokens and a mini-batches of size 256.
We use learning rate of 1e-4 and epoch of 30.

When training the student model with selected  pseudo labeled ASR outputs, most model hyperparameters are the same as in the grammaticality judgment model, with the exception of the batch size, learning rate, and number of training epochs.
We use batch size of 128, learning rate of 2e-5, and epoch of 10.

\begin{table}[t]\small
	\begin{center}
		\renewcommand{\arraystretch}{1.3}
		\begin{tabular}{l|ccc}
			\hline
			\bf    Method & \bf P & \bf R & \bf F1 \\
			\hline \hline
			
			
			Transition-based & 91.9 & 85.1 & 88.4\\
			BERT-Base  fine-tuning & 92.2 & 89.8 & 90.9\\
			ELECTRA-Small fine-tuning & 91.6 & 89.5 & 90.5\\
			ELECTRA-Base fine-tuning & 92.9 & 91.2 & 92.0\\
			Teacher fine-tuning & 92.5 & 92.1 & \textbf{92.3}\\
			\hline
			Unsupervised teacher & 86.8 & 62.0 & 72.3\\
			Our unsupervised &  90.2 & 89.1  & \textbf{89.6}\\
			\hline
		\end{tabular}
	\end{center}
	\caption{  Experiment  results on the Switchboard dev set. `` $*$ fine-tuning" means `` fine-tuning  $*$ model" on the Switchboard train set. The first part (from row 1 to row 5) is the supervised method using complicated hand-crafted features or contextualized word embeddings (e.g. ELMo \citep{peters-etal-2018-deep} and ELECTRA),  the second part (row 6 to 7) is the unsupervised methods. }
	\label{main result}
\end{table}

\begin{table}[t]\small
	\setlength{\tabcolsep}{4.7pt}
	\centering
	\small
	\renewcommand{\arraystretch}{1.3}
	\begin{tabular}{l|ccc}
		\hline
		\bf Method&\bf P&\bf R&\bf F1\\
		\hline\hline
		UBT \cite{wu-EtAl:2015:ACL-IJCNLP} &  90.3 & 80.5 & 85.1\\ 
		Bi-LSTM \citep{zayats2016disfluency} &  91.8 & 80.6 & 85.9\\ 
		NCM \citep{lou2018disfluency}  &  - & - & 86.8\\
		Transition-based \citep{wang2017transition}  &  91.1 & 84.1 & 87.5\\ 
		\hline
		Self-supervised\cite{wang2019multi} &93.4&87.3&90.2\\
		Self-training\cite{jamshid2020improving} &87.5&93.8&90.6\\
		EGBC\cite{bach2019noisy} &95.7&88.3& \bf91.8\\
		\hline
		\bf Our Method& 88.2  & 87.8 &\bf 88.0 \\
		\hline
	\end{tabular}
	\setlength{\abovecaptionskip}{4pt}
	\caption{Comparison with  previous state-of-the-art methods on the Switchboard test set. The first part (from row 1 to row 4) is the methods without using contextualized word embeddings (e.g. ELMo \citep{peters-etal-2018-deep} and ELECTRA),  the second part (row 5 to 7) is the methods  using contextualized word embeddings.}
	\label{compare-previous-work}
	\vspace{-1em}
\end{table}

\subsection{ Performance on English Switchboard }

As shown in Table \ref{main result}, we build six  baseline systems: (1) \textbf{Transition-based} is a neural transition-based model \citep{wang2017transition}. 
We directly use the code released by \citet{wang2017transition};\footnote{https://github.com/hitwsl/transition\_disfluency} 
(2) \textbf{BERT-Base  fine-tuning} means fine-tuning  BERT-Base model  on Switchboard train set;  
(3) \textbf{ELECTRA-Small fine-tuning } means fine-tuning  ELECTRA-Small (the discriminator) model  on Switchboard train set;  
(4) \textbf{ELECTRA-Base fine-tuning } means fine-tuning  ELECTRA-Base (the discriminator) model  on Switchboard train set;  
(5) \textbf{Unsupervised teacher}  is the  teacher model as shown in Step 2 of Algorithm \ref{algo:blah}; 
(6) \textbf{Teacher  fine-tuning} means fine-tuning  unsupervised teacher model  on Switchboard train set.

Table \ref{main result} shows the overall performances of our model on the Switchboard dev set.
Our unsupervised model achieves almost 17  point improvements over the baseline unsupervised teacher model.
Even compared with supervised systems using full set of Switchboard training data and contextualized word embeddings, our unsupervised approach achieves competitive performance.

Finally, we compare our  unsupervised model  to state-of-the-art supervised and semi-supervised methods from the literature  on the Switchboard test set, which can be divided into the following two categories: the methods without using contextualized word embeddings, and  the methods using contextualized word embeddings.
Table \ref{compare-previous-work}  shows that our unsupervised model is competitive with recent models using full set of Switchboard training data.
In particular, our unsupervised model even achieves slightly improvement over  the supervised methods without using contextualized word embeddings, demonstrating the effectiveness of our unsupervised model.

\begin{table}[t]\small
	\setlength{\tabcolsep}{5pt}
	\begin{center}
		\renewcommand{\arraystretch}{1.3}
		\begin{tabular}{l|c|c|c}
			\hline
			\bf    Method & \bf CallHome & \bf SCOTUS & \bf FCIC \\
			\hline
			Unsupervised teacher & 45.7 & 63.9 & 43.2\\
			\hline
			ELECTRA-Base & 60.9 & 79.4 & 62.8\\
			\hline
			Teacher fine-tuning & 63.7 & 81.9 & 64.3\\
			\hline
			Pattern-match &  65.2 & 79.9  & 66.1\\
			\hline
			Our unsupervised &  60.2 & 80.3 & 63.3\\
			\hline
		\end{tabular}
	\end{center}
	\caption{\label{re_error}   F1 scores on cross-domain disfluency detection. ``Pattern-match" \citep{zayats2018robust} is a  pattern match neural network architecture trained  on Switchboard train set, and  achieves state-of-the-art performance in cross-domain scenarios.}
	\label{cross_domain_result}
\end{table}

\begin{figure*}[!th]
	\centering
	\begin{minipage}{0.32\linewidth}\centering
		\includegraphics[width=5.5cm]{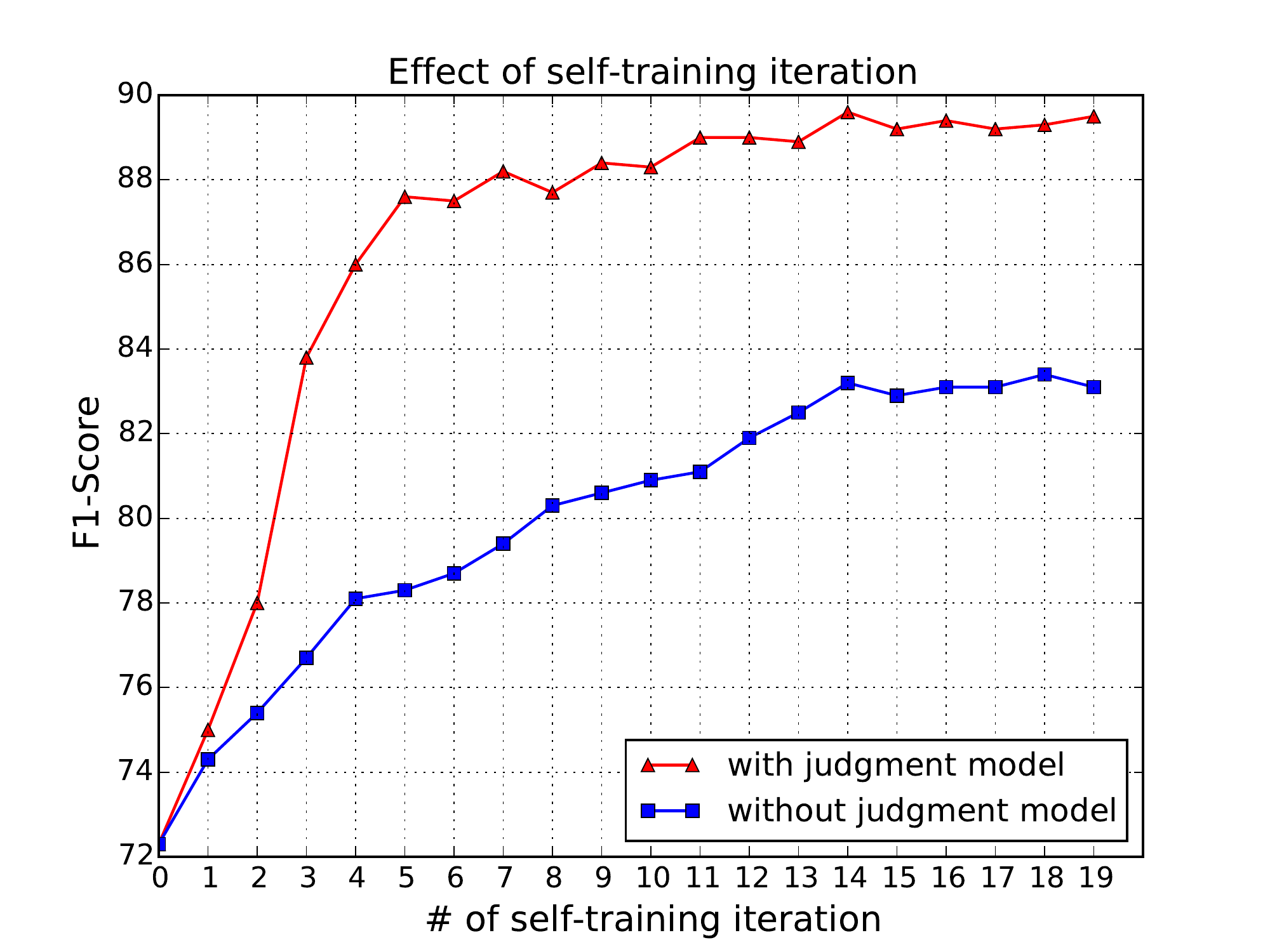}\\
		(a) 
	\end{minipage}
	\begin{minipage}{0.32\linewidth}\centering
		\includegraphics[width=5.5cm]{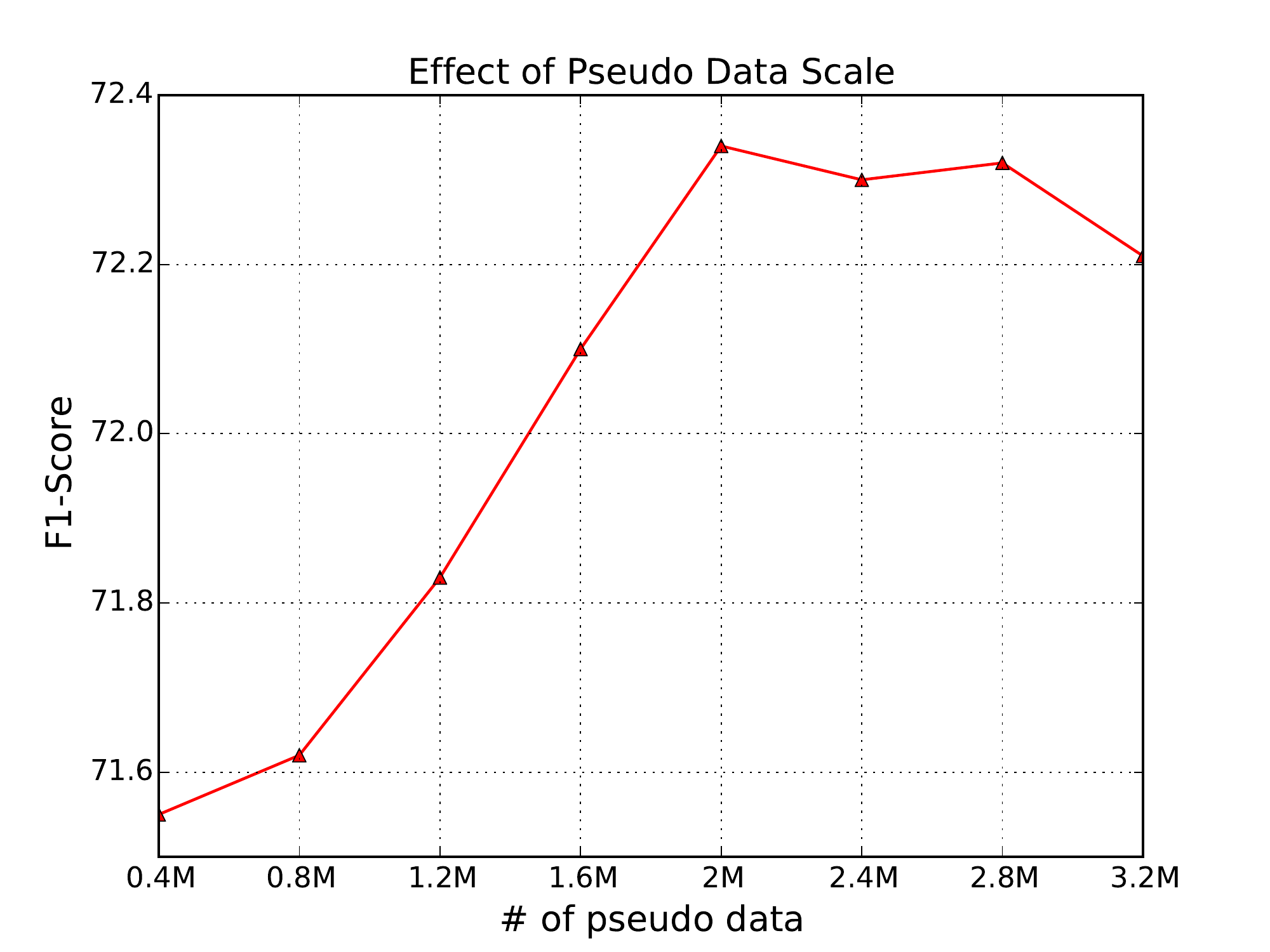}\\
		(b) 
	\end{minipage}
	\begin{minipage}{0.32\linewidth}\centering
		\includegraphics[width=5.5cm]{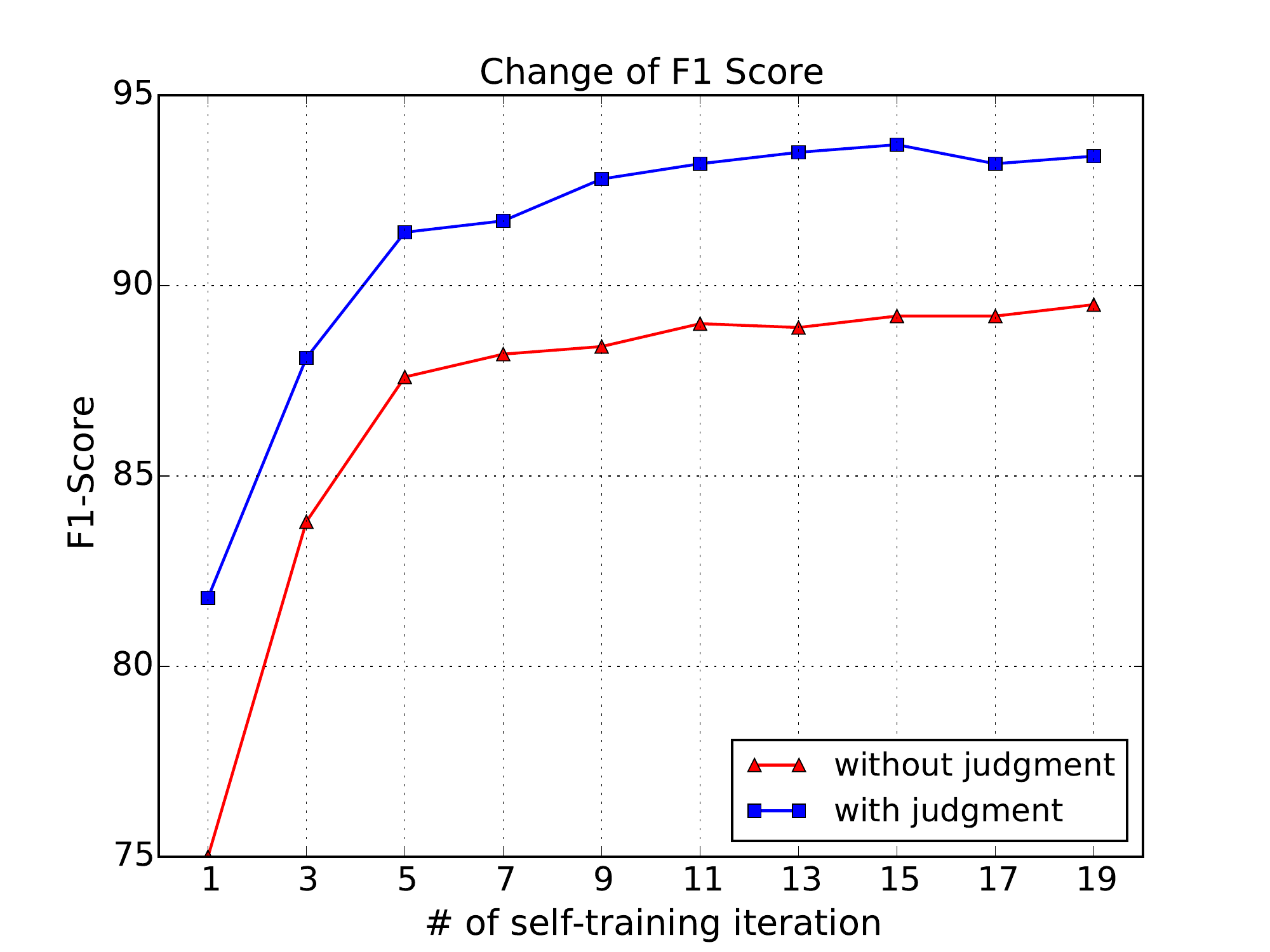}\\
		(c) 
	\end{minipage}
	\vspace{-0.0in}
	\caption{(\textbf{a}) Plot showing the effects of iterative training. (\textbf{b}) Plot showing the impact of pseudo training data size to the teacher model. (\textbf{c}) Plot showing the change of F1 score before and after using grammaticality judgment model. }
	\label{zhexian}
	\vspace{-0.0in}
\end{figure*}

\subsection{ Performance on Cross-domain Data }
To prove the robustness of our methods,  we also test our unsupervised model  on three out-of-domain publicly available datasets.
As shown in Table \ref{cross_domain_result}, we use four baseline systems: (1) \textbf{Unsupervised teacher}  is the  teacher model as shown in Step 2 of Algorithm \ref{algo:blah} trained  on pseudo train set; 
(2) \textbf{ELECTRA-Base fine-tuning } means fine-tuning  ELECTRA-Base (the discriminator) model  on Switchboard train set;  
(3) \textbf{Teacher fine-tuning } means fine-tuning unsupervised teacher model  on Switchboard train set;  
(4) \textbf{Pattern-match} \citep{zayats2018robust} means a  pattern match neural network architecture trained  on Switchboard train set, and  achieves state-of-the-art performance in cross-domain scenarios. 

For both the baseline and our unsupervised systems, we  directly use the model achieving state-of-the-art F1 score on the Switchboard dev set and directly test it on the out-of-domain data without retraining.
Table \ref{cross_domain_result} shows that our unsupervised model achieves consistent performance in both Switchboard and the three cross-domain datas.
In contrast to the performance on the Switchboard dev set as shown in Table \ref{main result}, our unsupervised model achieves performance similar to the ELECTRA-Base fine-tuning model.  
This surprising observation shows that our unsupervised model is  robust  in cross-domain testing.
We conjecture that our method uses a large amount of unlabeled  news data and ASR outputs, which make it survive the domain mismatch problem in cross-domain testing. 
Even compared with  the supervised Pattern-match model \citep{zayats2018robust}  achieving state-of-the-art performance in cross-domain scenarios, our model achieves competitive performance.

\begin{table}[t]\small
	\begin{center}
		\renewcommand{\arraystretch}{1.3}
		\begin{tabular}{l|c|c|c|c}
			\hline
			\bf    Method & \bf SWBD & \bf CallHome & \bf SCOTUS & \bf FCIC \\
			\hline
			Teacher &  72.3 & 45.7 & 63.9 & 43.2\\
			\hline
			No-select & 83.4 & 55.9 & 70.4 & 56.1\\
			\hline
			Select & 89.6 & 60.2 & 80.3 & 63.3\\
			\hline
		\end{tabular}
	\end{center}
	\caption{\label{ablation_test}   Ablation study of grammaticality judgment model. 
		``Teacher" means unsupervised teacher models. ``No-select" means our unsupervised self-training method without grammaticality judgment model. ``Select" means our unsupervised self-training method with grammaticality judgment model. ``SWBD" means the Switchboard dev set.}
	\label{ablation_result}
\end{table}

\section{ Ablation Studies }  

In this section, we study the importance of  grammaticality judgment model and iterative training.

\subsection{ The Importance of  Grammaticality Judgment Model }

To demonstrate the effect of grammaticality judgment model, we further conduct an experiment without grammaticality judgment model.
As shown in Table \ref{ablation_test},  both of our two models  achieve significant improvement compared with the baseline unsupervised teacher model.
Higher performance is achieved  through the introduction of grammaticality judgment model.
We conjecture that  grammaticality judgment model can help filter out the sentence with false pseudo labels.

\subsection{ A Study of Iterative Training }  

Here, we show the detailed effects of iterative training. 
As mentioned in Section 2.1, we first train a weak disfluency detection model on large-scale pseudo  data and then use it as the teacher to train a student model. 
Then, we iterate this process by putting back the new student model as the teacher model.

We plot F1-score  with respect to the number of iteration for  the two models with and without  grammaticality judgment model.
As shown in Figure ~\ref{zhexian} (a),  both the two models keep increasing until reaching an experiment upper limit, and achieve significant improvement over the model  in the first iteration.
 These results indicate that iterative training is effective in producing increasingly better models.

\section{Analysis}

\subsection{Varying Amounts of  Pseudo Data for Teacher Model}
We observed the impact of pseudo training data size to the teacher model as shown in Step 2 of Algorithm \ref{algo:blah}.
Figure ~\ref{zhexian} (b) reports the results of adding varying amounts of pseudo training data to the self-supervised teacher model.
We observe that  F1-score on the Switchboard dev set  keeps growing  until reaching an  upper limit when the amount of pseudo data increases.
The upper limit is only about 72.3 F1-score,  which is much lower than the supervised methods.
We conjecture that the distribution of our pseudo data is different from the distribution of the gold disfluency detection data, which limits the performance of our teacher model on real data.
The result also shows that disfluencies in ASR outputs are  complex, and disfluency detection cannot be fully solved by pretraining on pseudo disfluency data.

\begin{table}[t]\small
	\vspace{-0.0em}
	\setlength{\tabcolsep}{7pt}
	
	\begin{center}
		\renewcommand{\arraystretch}{1.3}
		\begin{tabular}{l | c | c | c}
			\hline
			\bf    Method & \bf Repet & \bf Non-repet & \bf Either \\
			\hline
			ELECTRA-Base & 95.6 &	77.7 &	92.0\\ 
			\hline
			Unsupervised Teacher & 91.4	& 54.3	& 72.3\\ 
			\hline
			Our Unsupervised  & 94.1 &	74.6 & 89.6\\
			\hline
		\end{tabular}
	\end{center}
	\vspace{-0.8em}
	\caption{F1-score of different types of reparandums on English Switchboard dev data.}
	\label{repetion-test}
	\vspace{-0.8em}
\end{table}

\subsection{Quantitative Analysis of Grammaticality Judgment Model}

The ablation test demonstrates the effect of grammaticality judgment model.
To prove the conjecture that  grammaticality judgment model  help filter out the sentence with false pseudo labels, we make two quantitative analyses for grammaticality judgment model.

The first quantitative analysis gives the classification accuracy  of grammaticality judgment model on the Switchboard dev set.
Grammaticality judgment model achieves a 85\% accuracy.
The result shows that grammaticality judgment model has the ability to judge whether an input  sentence is  grammatically-correct, and will always help select sentences with high-quality pseudo labels.

For the second quantitative analysis, we  observed the change of F1 score  by simulating the infer and select process of  iterative training on the Switchboard dev set.
For each iteration, we first use the teacher model to generate pseudo labels on the Switchboard dev set, and compute one  F1 score. 
Then we  use  grammaticality judgment model  to  select sentences.
We compute another F1 score on the selected sentences.
Figure ~\ref{zhexian} (c) reports the change of F1 score in each iteration.
The F1 score on selected sentences is always significantly higher than that without selecting.
The result shows that grammaticality judgment model can always help select sentences with high-quality pseudo labels.

\subsection{Repetitions vs Non-repetitions}

Repetition disfluencies are much easier to detect than other disfluencies, although not trivial since some repetitions can be fluent.
In order to better understand model performances, we evaluate our model's  ability  to detect repetition vs. non-repetition (other) reparandum on the Switchboard dev set. 
The results are shown in Table \ref{repetion-test}. 
All three models achieve high scores on repetition reparandum.
Our unsupervised model  is much better in predicting non-repetitions compared to the unsupervised teacher model. 
Even compared with  the supervised ELECTRA-Base model, our model achieves competitive performance on non-repetitions.
The result shows that our unsupervised model has the ability to  solve complex disfluencies.
We conjecture that our self-supervised tasks can capture more sentence-level structural information.

\section{Related Work}

\subsection*{Disfluency Detection}

Most work on disfluency detection focus on supervised learning methods, which mainly fall into three main categories: sequence tagging, noisy-channel, and parsing-based approaches. 
Sequence tagging approaches label words as fluent or disfluent using a variety of different techniques, including conditional random fields (CRF) ~\citep{georgila2009using,ostendorf2013sequential,zayats2014multi}, Max-Margin Markov Networks (M$^3$N) \citep{qian2013disfluency}, Semi-Markov CRF \citep{ferguson-durrett-klein:2015:NAACL-HLT}, and recurrent neural networks~\citep{hough2015recurrent, zayats2016disfluency,wang-che-liu:2016:COLING}. 
The main benefit of sequential models is the ability to capture long-term relationships between reparandum and repairs. 
Noisy channel models ~\citep{charniak2001edit,johnson2004tag,zwarts2010detecting,lou2018disfluency} use the similarity between reparandum and repair as an indicator of disfluency.
Parsing-based approaches \citep{rasooli2013joint,honnibal2014joint,wu-EtAl:2015:ACL-IJCNLP,yoshikawa2016joint, jamshid2019neural} jointly perform parsing and disfluency detection.
The joint models can capture long-range dependency of disfluencies as well as chunk-level information.
However, training a parsing-based model requires large annotated tree-banks that contain both disfluencies and syntactic structures.

All of the above works  heavily rely on human-annotated data.
There exist a limited effort to tackle the training data bottleneck.
\citet{C18-1299} and \citet{dong2019adapting}  use an autoencoder method to help for disfluency detection by jointly training the autoencoder model and disfluency detection model.
 \citet{wang2019multi} use self-supervised learning to tackle the training data bottleneck.
  Their self-supervised method can substantially reduce the need for human-annotated training data. 
  \citet{jamshid2020improving} shows that self-training and ensembling are effective methods for improving disfluency detection. 
  These semi-supervised methods achieve higher performance by introducing pseudo training sentences.
  However, the performance still  relies on human-annotated data.
  We explore  unsupervised disfluency detection, taking inspiration from the success of self-supervised learning and self-training on disfluency detection.

\subsection*{Self-Supervised Representation Learning}

Self-supervised learning aims to train a network on an auxiliary task where ground-truth is obtained automatically.
Over the last few years, many self-supervised tasks have been introduced in image processing domain, which make use of non-visual signals, intrinsically correlated to the image, as a form to supervise visual feature learning \citep{agrawal2015learning,wang2015unsupervised,doersch2015unsupervised}.

In natural language processing domain,  self-supervised research mainly focus on word embedding \citep{mikolov2013efficient,mikolov2013distributed} and language model learning \citep{bengio2003neural,peters-etal-2018-deep,radford2018improving}.
For word embedding learning, the idea is to train a model that maps each word to a feature vector, such that it is easy to predict the words in the context given the vector. 
This converts an apparently unsupervised problem  into a ``self-supervised" one: learning a function from a given word to the words surrounding it. 

Language model pre-training \citep{bengio2003neural,peters-etal-2018-deep,radford2018improving, devlin2018bert} is another line of self-supervised learning task. 
A trained language model  learns a function to predict the likelihood of occurrence of a word based on the  surrounding sequence of words used in the text. 
There are mainly two existing strategies for applying pre-trained language representations to down-stream tasks: feature-based and fine-tuning. 
The feature-based approach, such as ELMo \citep{peters-etal-2018-deep}, uses task-specific architectures that include the pre-trained representations as additional features. 
The fine-tuning approach, such as the Generative Pre-trained Transformer (OpenAI GPT)  \citep{radford2018improving} and BERT \citep{devlin2018bert}, introduces minimal task-specific parameters and is trained on the downstream tasks by simply fine-tuning the pre-trained parameters.

Motivated by the success of self-supervised learning,  we use self-supervised learning method to train a  weak disfluency detection model as teacher model.
We also  train a  sentence grammaticality judgment model  to help select sentences with high-quality pseudo labels.

\subsection*{Self-Training}

Self-training \citep{mcclosky2006effective} first uses labeled data to train a good teacher model, then use the teacher model to label unlabeled data and finally use the labeled data and unlabeled data to jointly train a student model.
Self-training  has also been shown to work well for a variety of  tasks including leveraging noisy data \citep{veit2017learning}, semantic segmentation \citep{babakhin2019semi}, text classification \citep{li2019learning}. 
\citet{xie2019self} present Noisy Student Training, which extends the idea of self-training  with the use of equal-or-larger student models and noise added to the student during learning. 

Our model builds upon the recent work on Noisy Student Training \citep{xie2019self} and further extend it to  unsupervised disfluency detection by  combining self-training and self-supervised learning methods.

\section{Conclusion}
	In this work, we explore  unsupervised disfluency detection by combining self-training and self-supervised learning. 
	We showed that it is possible to completely remove the need of human-annotated  data and train a  high-performance disfluency detection system in a completely unsupervised manner.

\section*{ Acknowledgments}
We thank the anonymous reviewers for their valuable comments.
This work was supported by the National Natural Science Foundation of China (NSFC) via grant 61976072, 61632011 and 61772153. 
Wanxiang Che is the corresponding author.

\bibliographystyle{acl_natbib} 
\bibliography{emnlp2020}

\end{document}